%% file: main.tex
\documentclass{article}

     \PassOptionsToPackage{numbers, compress}{natbib}

\usepackage[preprint]{neurips_2024}

\title{Understanding the dynamics of the frequency bias in neural networks}

\author{
Juan Molina \And  Mircea Petrache \AND  Francisco Sahli Costabal\And Mat\'ias Courdurier\AND
Pontificia Universidad Cat\'olica de Chile\\
\texttt{\{jjmolina1,mpetrache,fsahli1,mcourdurier\}@uc.cl}
}

\input{preamble}

\begin{document}

\maketitle

\begin{abstract} 
    Recent works have shown that traditional Neural Network (NN) architectures display a marked frequency bias in the learning process. Namely, the NN first learns the low-frequency features before learning the high-frequency ones. In this study, we rigorously develop a partial differential equation (PDE) that unravels the frequency dynamics of the error for a 2-layer NN in the Neural Tangent Kernel regime. Furthermore, using this insight, we explicitly demonstrate how an appropriate choice of distributions for the initialization weights can 
    eliminate or control the frequency bias. We focus our study on the Fourier Features model, an NN where the first layer has sine and cosine activation functions, with frequencies sampled from a prescribed distribution. In this setup, we experimentally validate our theoretical results and compare the NN dynamics to the solution of the PDE using the finite element method. Finally, we empirically show that the same principle extends to multi-layer NNs.
\end{abstract}

\section{Introduction}
An interesting open question for Deep Learning (DL) models is to find an informative way to correctly describe their learning dynamics. In the absence of well documented ad-hoc training strategies, practitioners are obliged to use architecture search heuristics and empirical hyperparameter tuning, to improve the predictive accuracy of Neural Networks (NNs), especially for finer (high-frequency) details in the data. In view of both cost-efficiency and environmental impact of DL training, there is an increasing need for a theoretical understanding of learning dynamics, which may lead to concise practices to improve the accuracy and efficiency in training of NNs. So far, multiple phenomena regarding the learning process have been studied: 

\textbf{Simplicity Bias.}
In DL models, it has been observed that the ``simplest'' features tend to be learned at earlier stages than more complicated features. This is the so-called \emph{simplicity bias} of DL models. The phenomenon has been first identified in basic architectures to frame different blunders for efficient learning \cite{arpit2017closer, geirhos2018imagenet, morwani2024simplicity, shah2020pitfalls,   yang2024identifying}, and beyond DL, being related to Occam's razor principle, it has also been proposed as a plausible structural element of generic learning processes \cite{ dingle2018input, feldman2016simplicity, schmidhuber1994discovering, shah2020pitfalls}. In DL, this phenomenon has been connected to the regularization aspect of stochastic gradient descent \cite{kalimeris2019sgd}, and to the subdivision in two phases of memorization/fine-tuning in the so-called Information Bottleneck Principle \cite{dahlgaard2016tight, tishby2015deep}.

\textbf{Spectral Bias / Frequency Principle.} 
A simple yet flexible way to quantify the simplicity bias is the so-called \emph{Frequency Principle} \cite{luo2019theory, xu2018understanding}, also known as \emph{Spectral (or Frequency) Bias} \cite{rahaman2019spectral}, which refers to the experimentally observed tendency of NNs to learn lower frequency features earlier than higher ones. This phenomenon has been empirically proven in studies such as \cite{xu2019frequency, xu2019training}. Furthermore, the existence of frequency bias has been investigated using various techniques. For instance, shallow NNs with $\tanh$ activation functions were analyzed in \cite{xu2018understanding}, while \cite{cao2021towards, ronen2019convergence} explored frequency bias through the lens of eigenvalues theory and certain data hypotheses. Additionally, \cite{ramasinghe2022frequency} provides a qualitative approach to avoiding frequency bias by assuming the existence of optimal parameters. Similarly, \cite{luo2022upper, luo2022exact,  zhang2021linear} gives clues on how to mitigate frequency bias under the infinite-width condition of hidden layers. The theory behind the frequency bias has been applied to enhance physics-informed NNs (PINNs) \cite{wang2021eigenvector}, to study overfitting \cite{xu2021deep, zhang2021linear}, and to adjust gradient descent methods \cite{geifman2022spectral, geifman2023controlling}.

\textbf{Fourier Features.} Implicit Neural Representations, a.k.a. Neural Fields, are an emerging research area where multiple techniques to mitigate the frequency bias have been developed. The goal of these architectures is to accurately approximate continuous signals $f : \R^d \to \R^l$, such as images, videos, complex 3D shapes, sensor measurements, etc., with fully connected NNs for tasks such as reconstruction, superresolution, editing, etc. \cite{molaei2023implicit, xie2022neural}. Given that many of the signals of interest in this field present high-frequency components that should be learned by the NNs, several empirical solutions have been proposed to alleviate the frequency bias. For instance, using sinusoidal or Gaussian activation functions, allowed increased accuracy in multiple applications \cite{ramasinghe2022beyond, sitzmann2020implicit}. Fourier features (FF), an ad-hoc type of positional encoding, is one of the most successful approaches to control the frequency bias in Neural Fields \cite{tancik2020fourier}. Here, an input coordinate is preprocessed in the first layer by projecting it onto a high-dimensional vector consisting of cosine and sine of random frequencies. This idea was previously introduced and utilized in the field of kernel machines \cite{rahimi2007random, sinha2016learning}, transformers \cite{vaswani2017attention}, and more general architectures \cite{benbarka2022seeing, ramasinghe2021learning,zheng2021rethinking}.

\textbf{Neural Tankent Kernel regime.} Despite its success, there is limited theoretical understanding on how the hyperparameters of FF affect the frequency bias and the training process. In this work, we study the FF architecture to give a rigorous description of frequency bias dynamics, by working in the Neural Tangent Kernel regime (NTK), in which the learning dynamics are studied in the infinite-width limit of NNs layers \cite{arora2019exact, jacot2018neural, lee2019wide}. The NTK is one of the most prominent tools to study learning process in NNs, as it allows to obtain a deterministic kernel representation of the training error, leading to a natural tool to analyze frequency bias \cite{tancik2020fourier}.

\textbf{Related Work.} The NTK for FF architectures has been used as an initial formalization of the model \cite{tancik2020fourier}. Successively, NTK has been used to study the frequency principle in recent works \cite{luo2022exact, yang2019fine}, including approaches using replica symmetry \cite{bordelon2020spectrum}. The latter thesis also includes a study of frequency bias via eigenvalues of the NTK kernel, including some aspects of the dynamics. Heuristics about the influence of activation functions over frequency bias (without using NTK) follow from computations in \cite{hong2022activation}, and the influence towards spectral bias of initialization choices under stochastic gradient descent is discussed in \cite{macdonald2023progressive}. However, none of these previous studies provides a detailed enough result on the dynamics of the frequency for a FF architecture, in order to directly compute the influence of initialization distribution on the dynamics.

\subsection*{Main contributions}
\begin{itemize}
\item Based on the NTK, we give a rigorous derivation 
of the Partial Differential Equation (PDE) describing the frequency dynamics of errors of a FF neural network with one hidden layer. 
\item Based on the above equation we provide a novel understanding about the precise influence on Frequency Bias of the initialization distributions of the weights. 
\item Experimentally, we verify that a Finite Element simulation of our model's PDEs qualitatively reproduces the spectral bias of a NN. We also show that our results still hold for multilayer NNs.
\end{itemize}

\section{Problem setup}

In this section, we introduce our NN model and the dynamics that follows during training in order to approximate a target function.

Let $\widetilde{f}:\mathbb{R}^{d}\rightarrow\mathbb{R}$ be our target function such that $\widetilde{f}\in C\left(\mathbb{R}^{d}\right)\cap L^{2}\left(\mathbb{R}^{d}\right)$.
Our neural network is defined via a parameterized approximate of $\widetilde{f}$, denoted $f(\boldsymbol{x},\Theta(t)):\mathbb R^d\times \mathbb R^{(d+1)\times m}\to \mathbb R$ depending on parameter $\Theta:[0,T]\to\mathbb R^{(d+1)\times m}$. We take this parameterization to be of the following specific form, depending on a nonlinear point-wise activation function $g:\mathbb{R} \to\mathbb R$:
\begin{equation}
f(\boldsymbol{x},\Theta(t)) = \frac{1}{\sqrt{m}}\sum_{k=1}^{m} a_{k}(t)g\left(\bsw_{k}(t)\cdot \bsx\right), \label{eq:NNfirst}
\end{equation}
where $\bsa (t) =(a_1(t),\dots,a_m(t))\in \R^{m}$, $\mathbf W(t)=(\bsw_1(t), ..., \bsw_m(t)) \in R^{d\times m}$ and $\Theta(t)=\{\bsa (t), W(t)\}$.  

To train the NN, we consider square error loss $\mathcal{L}(\bsx,f):=\frac12(f(\bsx,\Theta(t)) - \widetilde{f}(\bsx))^2$, and to model stochastic gradient descent we average it over random i.i.d. batches $S:=\left\{\bsx_i\right\}_{i=1}^N$ from the training dataset, i.e. we minimize the empirical risk
\begin{alignat}{1}
\mathcal R_S(\Theta) & :=\frac1N\sum_{i=1}^N\mathcal L(\bsx_i,f)=\frac{1}{2N}\sum_{i=1}^{N}\left(f(\bsx_i,\Theta)-\widetilde{f}(\bsx_{i})\right)^{2}. \nonumber
\end{alignat}

The dynamics of the parameter $\Theta$ is approximated by continuous gradient descent of the empirical risk $\mathcal R_S$, that is, 
\begin{alignat}{1}
\frac{d}{dt}\Theta(t) & =-\nabla_{\Theta}\mathcal R_S(\Theta(t))=-\frac1N\sum_{i=1}^{N}\nabla_{\Theta}f(\bsx_{i},\Theta(t))\left(f(\bsx_i,\Theta(t))-\widetilde{f}(\bsx_{i})\right), \nonumber
\end{alignat}
and the evolution of the NN is modeled by the solution of the ordinary differential equation:
\begin{alignat}{1}\label{eq:initialdynamics}
 \frac{d}{dt}f(\bsx,\Theta(t)) 
 & =\nabla_{\Theta}f(\bsx,\Theta(t))\cdot\frac{d}{dt}\Theta(t) \\ 
 & =-\frac1N\sum_{i=1}^{N}\nabla_{\Theta}f(\bsx,\Theta(t))\cdot\nabla_{\Theta} f(\bsx_{i},\Theta(t))\left(f(\bsx_i,\Theta(t))-\widetilde{f}(\bsx_{i})\right) \nonumber\\ 
 & =-\frac1N\sum_{i=1}^{N}K_m(\bsx,\bsx_{i}; t)\left(f(\bsx_i,\Theta(t))-\widetilde{f}(\bsx_{i})\right), \nonumber
\end{alignat}
where we introduced a time-dependent kernel $K_m:\mathbb R^d\times\mathbb R^d\times[0,T]\to\mathbb R$, defined as follows (with notation as in \eqref{eq:NNfirst})
\begin{alignat}{1}\label{eq:initialkernel}
K_m(\bsx,\bsxp; t):=&\nabla_{\Theta} f(\bsx,\Theta)\cdot\nabla_{\Theta} f(\bsxp,\Theta)\\
=&\sum_{k=1}^m\frac{\partial f}{\partial a_{k}}(\bsx,\Theta)\cdot\frac{\partial f}{\partial a_{k}}(\bsxp,\Theta)+ \sum_{k=1}^m\frac{\partial f}{\partial \bsw_{k}}(\bsx,\Theta)\cdot\frac{\partial f}{\partial \bsw_{k}}(\bsxp,\Theta) \nonumber
\end{alignat}

\subsection{Assumption: Infinite width regime}\label{assu}
In this work, we study the training dynamics under the NTK regime \cite{arora2019exact, jacot2018neural, lee2019wide}. This means with have two assumptions: 
  
  \textbf{(i)} The initialization weights $\{\bsw_{k}(0)\}_{k=1}^m$ are i.i.d. random variables drawn from $\rho_{\bsw}$ with zero mean and covariance matrix $\mathsf{Var}[\bsw(0)]=\sigma_{\bsw}^2I_{d\times d}$, and the parameters $\{a_{k}(0)\}_{k=1}^m$ are taken to be i.i.d. random variables drawn from the distribution $\rho_a$ with mean zero and finite variance $\sigma_a^2$. Note that the normalization is nonstandard from a probabilistic point of view.

 \textbf{(ii)} The hidden layer dimension $m$ tends to infinity. 
 
 Then the kernel $K_m(\bsx,\bsxp; t)$ from \eqref{eq:initialkernel} can be approximated with high probability by the continuum kernel $K$ defined as:
\begin{equation}\label{eq:NTKkernel}
    K(\bsx, \bsxp) := \mathbb{E}_{\theta\sim\rho_a\times\rho_{\bsw}}\left[\nabla_{\theta} f^0(\bsx,\theta)\cdot\nabla_{\theta} f^0(\bsxp,\theta)\right]
\end{equation}
where $f^0(\bsx,\theta):=a\ g(\bsw\cdot \bsx)$ for $\theta=(a,\bsw)$. In other words, NTK theory says that 
\begin{equation}
    K_m\left(\bsx,\bsxp; t\right)\stackrel{(m\to\infty)}{\simeq}  K(\bsx, \bsxp).
\end{equation}
Note that the whole NTK evolution is determined by $\rho_a\times\rho_{\bsw}$, the distribution from which $\theta_k=(a_k,\bsw_k), k=1,\dots,m$ are i.i.d.-sampled at initialization.

From now on, we work with the limit ``tangent'' kernel $K$ from \eqref{eq:NTKkernel} instead of $K_m$. In particular, note that $K$ only depends on the initialization distributions $\rho_a, \rho_{\bsw}$ and the activation function $g$. 

\subsection{Dynamics of the error}

Summarizing, under these assumptions we can approximate the dynamics of the NN by:
\begin{equation}
 \frac{d}{dt}\left(f(\bsx,\Theta(t))-\widetilde{f}(\bsx)\right) = -\frac1N\sum_{i=1}^N K(\bsx,\bsx_{i})\left(f(\bsx_i,\Theta(t))-\widetilde{f}(\bsx_{i})\right). 
\end{equation}
that can be conveniently approximated using the Central Limit Theorem by:
\begin{equation}\label{eq:dinamicalinealespacioimagen}
    \frac{du}{dt}(\bsx,\Theta (t))= - \int_{\mathbb{R}^{d}}K\left(\bsx,\bsxp\right)u\left(\bsxp,\Theta (t)\right)d\rho_{\text{data}}(\bsxp),
\end{equation}
where $u(\bsx,\Theta(t))=f(\bsx,\Theta(t))-\widetilde{f}(\bsx)$ and $\rho_{{\text{data}}}(\bsx)$ is the density of the data distribution.
This representation is called linearized residual dynamics, introduced in \cite{mei2019mean}.

\section{Derivation of the frequency bias equation}

Now, using equation (\ref{eq:dinamicalinealespacioimagen}) we will deduce the frequency dynamics. These Fourier computations are formulated in the framework of tempered distributions. Let $\mathcal{S}:=\mathcal{S}(\mathbb{R}^{d})$ be the Schwartz space on $\mathbb{R}^{d}$ and $\mathcal{S}':=\mathcal{S}'(\mathbb{R}^{d})$ be the space of tempered distributions. The distributional Fourier transform is defined as
\begin{equation}
    \left\langle\mathcal{F}[h],\phi\right\rangle =      
    \left\langle h,\mathcal{F} [\phi]\right\rangle
\end{equation}
where $h\in\mathcal{S}'$, $\phi\in\mathcal{S}$ and $\mathcal{F}$ is the usual Fourier transform defined as
\begin{equation}
    \widehat{h}(\bxi) = \mathcal{F}[h](\bxi) := \int_{\R^d} h(\bsx) e^{-2\pi i \bsx \cdot \bxi} d\bsx
\end{equation}
\begin{theorem}[proved as Thm. \ref{thm:evol_app}]\label{thm:evol}
Under the assumptions \ref{assu}, and assuming that $g \in H^{1}$, the dynamics (\ref{eq:dinamicalinealespacioimagen}) can be expressed in frequency space (in the sense of distributions, i.e. in duality with an arbitrary Schwarz test function $\psi\in \mathcal S(\mathbb R^d)$) as:
\begin{alignat}{1}
\left\langle \frac{d}{dt}\widehat{u},\psi\right\rangle 
= & -\frac{\sigma_{a}^{2}}{4\pi^{2}} \int_{\mathbb{R}^{d}}   \overline{\nabla\psi(\bxi)} \cdot \int_{\mathbb{R}^{2}} \nabla{\widehat{u}_{\rho_{\text{data}}}\left(\frac{y'}{y}\bxi\right)} \rho_{\bsw}\left(\frac{1}{y}\bxi\right)   \frac{\overline{\widehat{g'}\left(y'\right)} \widehat{g'}\left(y\right)}{\lvert y \rvert^d}dy'dy d\bxi \label{eq:dynamicsfinal}\\
 & -\int_{\mathbb{R}^{d}}   \overline{\psi(\bxi)} \int_{\mathbb{R}^{2}} {\widehat{u}_{\rho_{\text{data}}}\left(\frac{y'}{y}\bxi\right)} \rho_{\bsw}\left(\frac{ 1}{ y }\bxi\right)  \frac{\overline{\widehat{g}\left(y'\right)} \widehat{g}\left(y\right)}{\lvert y \rvert^d}dy'dy d\bxi . \nonumber
\end{alignat}
where $\hat{u}_{\rho_{\text{data}}}=\mathcal{F}[u\rho_{\text{data}}]$.
\end{theorem}

This initial result (\ref{eq:dynamicsfinal}) sheds light on how the dynamics of the error frequency $\widehat{u}$ depend on five aspects: the dimension $d$, the Fourier transform of the activation function $g$ and its derivative $g'$, the initial parameter distribution $\bsw$, and the variance of the parameter $a$. However, interpreting this equation (\ref{eq:dynamicsfinal}) proves challenging since it involves an integral operator combining the different frequencies through an integral. Thus, to give a more interpretable explanation of the frequency dynamics we apply this theorem to the specific case of FF.

\section{Dynamics of frequency bias for the Fourier features model}

\input{./plots/FFmodel}

To approximate the model of Fourier Features, we define our neural network
as 
\begin{equation}
f(\bsx,\Theta(t)) = \frac{1}{\sqrt{2m}}\sum_{k=1}^{m}a_{k}(t)g_1\left(\bsw_{k}(t)\cdot \bsx\right)+\frac{1}{\sqrt{2m}}\sum_{k=1}^{m}b_{k}(t)g_2\left(\bsw_{k}(t)\cdot \bsx\right)
\label{FFmodel}
\end{equation}
here $\Theta(t)=\{\bsa (t), \bsb (t), W(t)\}$, where $\bsa \in \R^{m}$, $\bsb \in \R^{m}$ and $\mathbf{W}(t)=(\bsw_1(t), ..., \bsw_m(t))^T \in \R^{d}\times \R^{m}$ are the trainable parameters with initial distribution $\rho_{a}$, $\rho_{b}$ and $\rho_{\bsw}$ respectively (See figure \ref{fig:FFmodel}). 
In this case, the same methodology and analysis as above can be made, where the only difference is the expression of kernel $K$, which is now given by 
\begin{alignat}{1}
K(\bsx,\bsxp)=  & \mathbb{E}_\theta\left[a^{2}g_{1}'\left(\bsw\cdot \bsx\right)g_{1}'\left(\bsw\cdot \bsxp\right)\bsx\cdot \bsxp\right]+ \mathbb{E}_\theta \left[b^{2}g_{2}'\left(\bsw\cdot \bsx\right)g_{2}'\left(\bsw\cdot \bsxp\right)\bsx\cdot \bsxp\right] \\
 & +\mathbb{E}_\theta\left[g_{1}\left(\bsw\cdot \bsx\right)g_{1}\left(\bsw\cdot \bsxp\right)\right]+ \mathbb{E}_\theta\left[g_{2}\left(\bsw\cdot \bsx\right)g_{2}\left(\bsw\cdot \bsxp\right)\right] \nonumber
\end{alignat}

\begin{corollary}[Proved as Cor. \ref{cor:PDE_app}]\label{cor:PDE}
Under the assumptions \ref{assu}. If the NN is given by equation (\ref{FFmodel}) where $g_1(x)=\cos(2\pi x)$, $g_2(x)=\sin(2\pi x)$, $\mathbb{E}[a^2]=\mathbb{E}[b^2]=\sigma_{a}^{2}$ and $\rho_{\bsw}$ is an even function, then dynamics (\ref{eq:dynamicsfinal}) in the frequency space is given by
\begin{equation}
\frac{d}{dt}\widehat{u}(\bxi,\Theta(t)) 
= \sigma_{a}^{2} \mathrm{div}\left(\rho_{\bsw}(\bxi)\nabla{\widehat{u}_{\rho_{\text{data}}}\left(\bxi, \Theta(t)\right)}  \right) - \rho_{\bsw}( \bxi){\widehat{u}_{\rho_{\text{data}}}\left(\bxi,\Theta(t)\right)}, \qquad \bxi \in \R^{d}, t>0,
\label{eq:dynamicsFF}
\end{equation}
with initial condition
\begin{equation}
    \widehat{u}_0(\bxi) = \widehat{u}(\bxi,\Theta(0)) = \mathcal{F}_{\bsx \rightarrow \bxi} \left[ f(\bsx,\Theta(0))-\widetilde{f}(\bsx) \right].
\end{equation}

\end{corollary}

\begin{remark}\label{remark:frozen}
 In the setting of a FF neural network with frozen random weight $\mathbf{W}(t)=\mathbf{W}(0)$, equation \eqref{eq:dynamicsFF} becomes
\begin{equation}
\frac{d}{dt}\widehat{u}(\bxi,\Theta(t)) 
= - \rho_{\bsw}( \bxi){\widehat{u}_{\rho_{\text{data}}}\left(\bxi,\Theta(t)\right)}.  
\label{eq:dynamicsFF_frozenW}
\end{equation}
which is equation \eqref{eq:dynamicsFF}  
with $\sigma_a=0$.
\end{remark}

From now on, we assume that the sampling is rich enough, and for the ensuing analysis we approximate  $\widehat{u}_{\rho_{\text{data}}}\approx \widehat{u}$
.

\subsection{Qualitative analysis of the frequency dynamics}

FF-specific equation (\ref{eq:dynamicsFF}) corresponds to a damped heat equation and therefore it provides a quite interpretable expression that is easier to understand. Indeed, the roles played by the initial distributions of the parameters are very explicit: the initial distribution $\rho_a$ only influences the dynamics through the value of its variance $\sigma_a^2$, and $\sigma_a^2$ only multiplies the magnitude of the diffusion term. The distribution $\rho_{\bsw}$ appears in the diffusion coefficient and as the magnitude of the damping term, hence, it is the only element introducing frequency dependency in the dynamics. Notably, from equation (\ref{eq:dynamicsFF}), we can also observe that {\em the input dimension $d$ does not play any role in any frequency dependent behaviour of the dynamics}.

\subsubsection{The role of the distribution of $\rho_{\bsw}$}

The form of equation \eqref{eq:dynamicsFF} confirms that the NN may exhibit a frequency bias during learning. For small $\sigma_a$ or frozen weights $\mathbf{W}$, the solution to equation \eqref{eq:dynamicsFF} is approximated by the solution of equation~\eqref{eq:dynamicsFF_frozenW}, which is explicitly:
\begin{equation}\label{eq:sigma=0}
\widehat{u}(\bxi,\Theta(t)) = \widehat{u}_0(\bxi) e^{-\rho_{\bsw}(\bxi) t}.
\end{equation}
For further analysis, we define the frequency learning rate $\kappa(\bxi)$ as the slope of the learning dynamics at the beginning of the training process, such that:
\begin{equation}
    \log | \widehat{u}(\bxi,\Theta(t)) | \approx -\kappa(\bxi)t +  \log | \widehat{u}_0(\bxi) |, \qquad \text{ for } t\sim0.
\end{equation}
Then, it stems from equation \eqref{eq:sigma=0}  that the learning rate at each frequency is exactly:
\begin{equation}\label{eq:learningrate}
\kappa(\bxi) = \rho_{\bsw}(\bxi).
\end{equation}
Now, if we aim to mitigate the frequency bias, it is necessary for all frequencies to be learned at the same rate. A first natural choice is to take $\rho_{\bsw}$ to be \emph{constant} across all frequencies. This corresponds to choose a uniform distribution, as done in \cite{sitzmann2020implicit}, that covers all the frequencies of the target function $\widetilde{f}$. Nonetheless, since the frequency content of $\widetilde{f}$ is not know a priory, we run the risk of leaving some frequencies of $\widetilde{f}$ outside of the learnable range. Therefore, it is more robust to choose $\rho_{\bsw}$ as a normal distribution with a large enough standard deviation, which is always positive. A disadvantage of choosing $\rho_{\bsw}$ as a
normal distribution with a very large standard deviation, is that the frequency learning rate is inversely proportional to the maximum value of $\rho_{\bsw}$. Which also arises as an issue when choosing a uniform distribution with a very large support. We will test these observations in the experiment section (see Figure \ref{fig:learningrate})

\subsubsection{The role of the variance $\sigma_a^2$}

The second form of control is indirectly determined by the distribution $\rho_a$. In this case, the distribution itself is not crucial: \emph{only its variance $\sigma_a^2$ appears in \eqref{eq:dynamicsFF}}. Furthermore, $\sigma_{a}^{2}$ impacts the diffusion term (first term in \eqref{eq:dynamicsFF}), and not the damping term.

In \cite{tancik2020fourier}, where Fourier Features are introduced, it is discussed that training the frequency parameters in the first layer does not yield any benefits, therefore it is preferable to leave this layer untrained. This observation is explained by our findings: the learning dynamics for frozen $\mathbf{W}(t)$ are practically the same as the dynamics in the case of $\sigma_a$ small, as is mentioned in the Remark \ref{remark:frozen}. Nonetheless, when $\sigma_a$ is large enough, the diffusive term in equation \eqref{eq:dynamicsFF} will play a relevant role in the learning dynamics, and there will be noticeable differences with the training under frozen $\mathbf{W}(t)$. In particular, the non-local nature of the diffusion term should help the NN to learn frequencies beyond the initial range allowed by $\rho_{\bsw}$. This claim is validated by empirical experimentation (see figure \ref{fig:FEMvsNN}).
A negative aspect of choosing a large $\sigma_a$ is that the initial estimation $f(x,\Theta(0))$ will be far from $\widetilde{f}$, therefore requiring more iterations for the NN to achieve a good estimation of $\widetilde{f}$.

\section{Numerical Experiments}

\begin{figure}[htb]
	\centering
		\includegraphics[width=.9\linewidth]{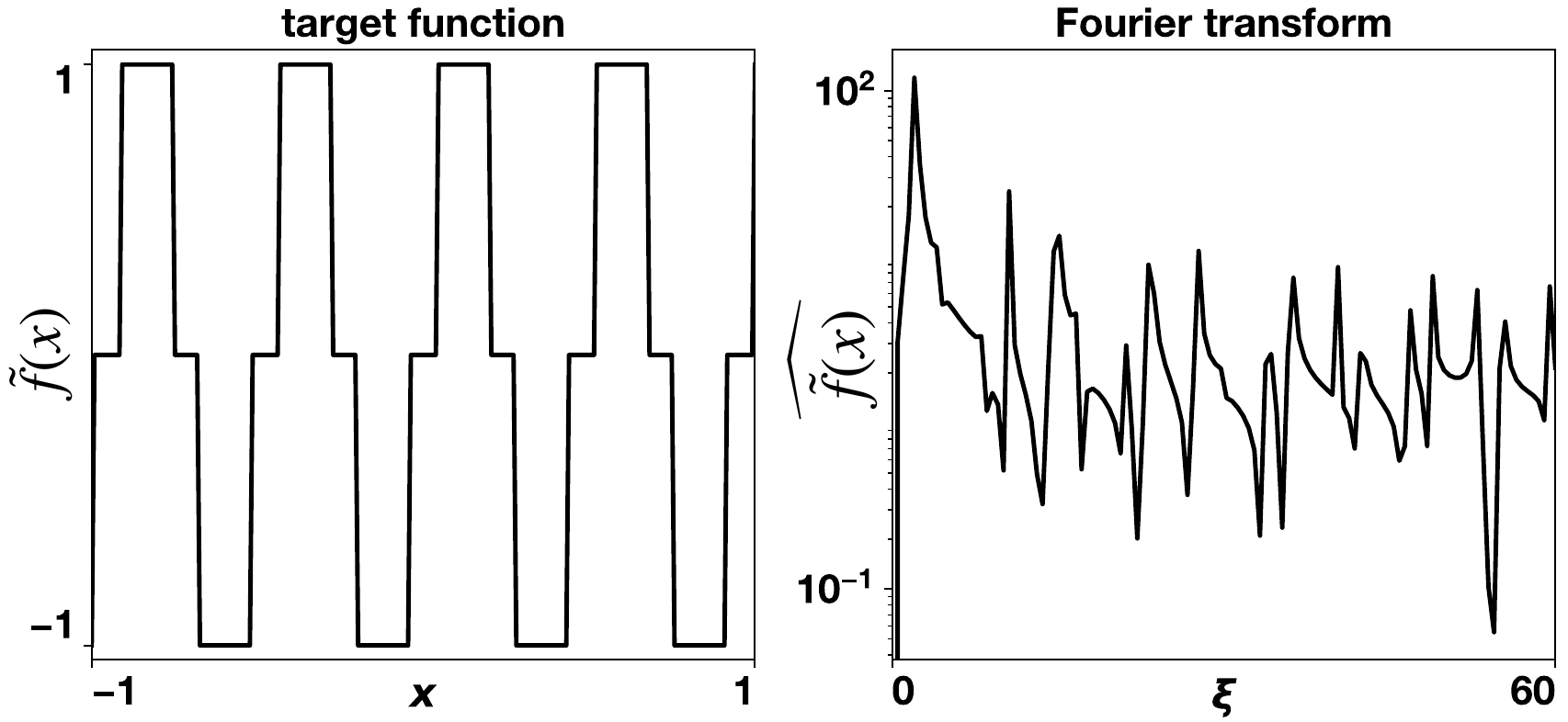}
	
	\caption{Target function $\widetilde{f}$ used in our experiments and the magnitude of its Fourier transform.}
    \label{fig:exampl}
\end{figure}

To validate our results in this section we analyze how the NNs defined in (\ref{FFmodel}) with $m=2000$ learns the target function given by $\widetilde{f}(x)=\text{round}(\sin((4.2)\pi x))$, where $\text{round}$ is the function that returns the closest integer to its argument (see Figure \ref{fig:exampl}).

\subsection{NN Training}
We generate a training set by evaluating the target function $\widetilde{f}(x)$ in an equispaced grid from -1 to 1 using 240 points. The initial parameters are sampled from $\rho_a \sim \mathcal{N}(0,\sigma_a)$ and $\rho_{\bsw} \sim \mathcal{N}(0,\sigma_{\bsw})$ or $\rho_{\bsw} \sim \mathcal{U}(-R,R)$. The value of $\sigma_a$ was $2/\sqrt{4000}$ (labeled as $0.03$). Additionally,  we train without batching the train set, and we use gradient descent with an optimization learning rate of $10^{-5}/240$. The optimization process is carried out until the 10000th iteration. For our experiments, we train with 100 random initializations and present the results as averages. We use JAX \cite{jax2018github} to implement and train the NN and FEniCS \cite{AlnaesEtal2015, LoggEtal2012} to solve the PDE in \eqref{eq:dynamicsFF} with the finite element method (FEM). More details regarding the implementation are presented in Appendix~\ref{FEMapprox}. In terms of hardware, we used a personal laptop with an NVIDIA GeForce GTX 3060 with 6GB memory.

\subsection{Frequency Learning Rates}

\begin{figure}[ht]
	\centering

		\includegraphics[width=\linewidth]{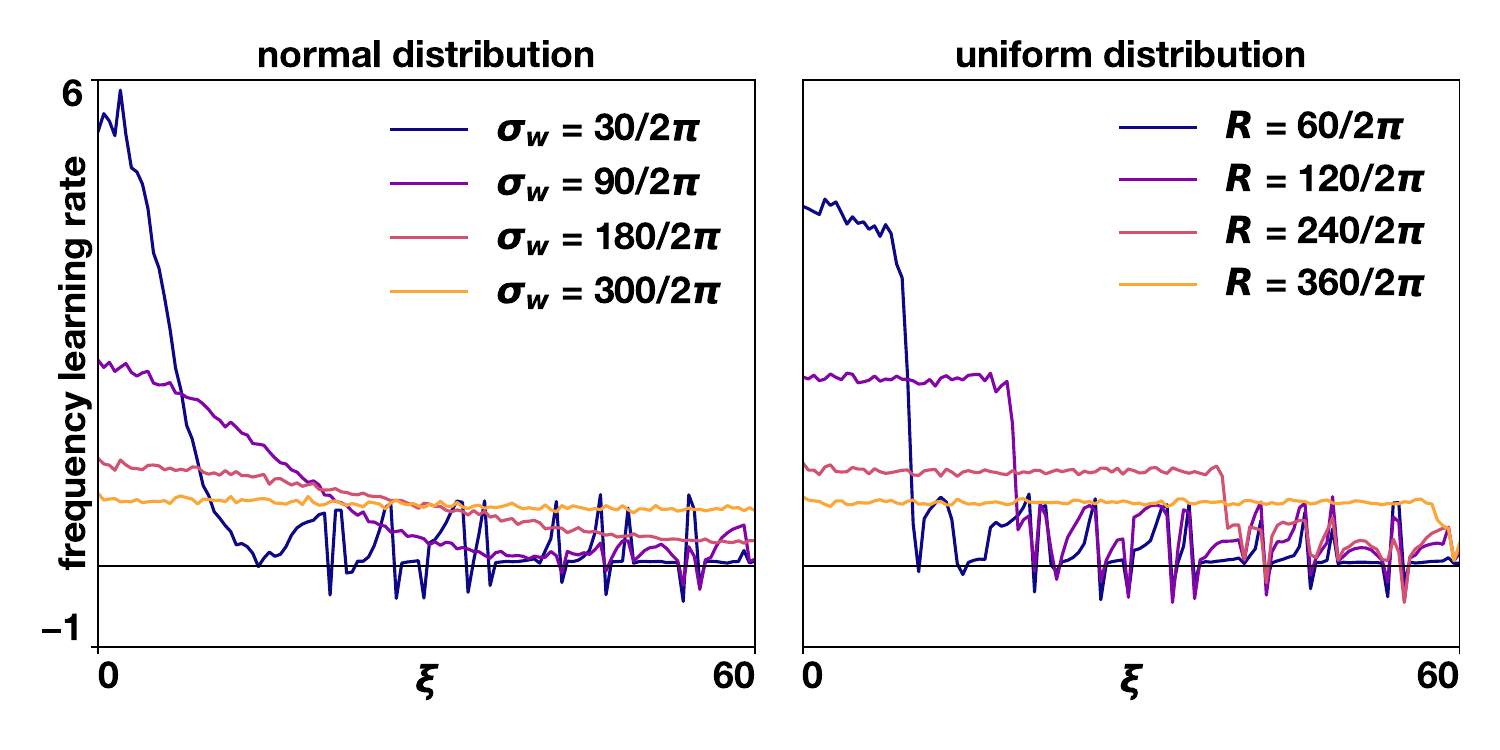}

	\caption{Frequency learning rate $\kappa(\bxi)$ for different initialization distributions $\rho_{\bsw}(\bxi)$. Left panel: normal distribution with different standard deviation $\sigma_{\bsw}$, right panel: uniform distribution with different widths $R$.}\label{fig:learningrate}
\end{figure}

For the first experiment, we employed the 100 trained NNs to calculate the frequency learning rate $\kappa(\bxi)$ defined in equation \eqref{eq:learningrate}. We conducted this analysis for both normal and uniform distributions of $\rho_{\bsw}$ with varying variances (refer to Figure \ref{fig:learningrate}).

Figure \ref{fig:learningrate} validates our results. Indeed, the distribution $\rho_{\bsw}$ is the primary variable for controlling or tuning the frequency bias. For instance, to eliminate spectral bias, the best options are to use a normal distribution with a standard deviation of $\sigma_{\bsw} = 300/2\pi$ or larger, or a uniform distribution with $R$ greater than or equal to 60. Furthermore, if the value of $\sigma_{\bsw}$ or $R$ is not large enough, the NN is incapable of learning all frequencies, as demonstrated in the cases with $\sigma_{\bsw} = 30/2\pi$ or $\sigma_{\bsw} = 90/2\pi$, or more evidently in the case of uniform distribution, where in each case, frequencies above the corresponding value of $R$ are not learned.

The structural noise (oscillations) shown in Figure \ref{fig:learningrate} on both distributions (in the case $\sigma_w = 30/2\pi$ and $90/2\pi$ on the Gaussian distribution and $R=60/2\pi,120/2\pi$ and $240/2\pi$ in the uniform distribution) is due to Discrete Fourier transform trying to approximate the last frequency in the domain of $\rho_{\bsw}$. This phenomenon is known as spectral leakage \cite{lyon2009discrete} and it is amplified due to the logarithmic scale. For instance, in the case of  $\sigma_{\bsw} = {30}/{2\pi}$ (normal distribution), the NN only learns approximately up to the frequency $\xi=15$. Similarly, in the case of uniform distribution, the blue line tells us that the NN is only capable of learning frequencies lower than the frequency $\xi = {60}/{2\pi} \approx 9.55$.

\subsection{Comparation NN dynamics vs FEM modeling}

We perform numerical simulations of the PDE \eqref{eq:dynamicsFF} to validate our results. Although Equation (\ref{eq:dynamicsFF}) is defined over all $\mathbb{R}$, for the numerical experiments we need to restrict the domain, to $\Omega = (-60, 60)$, and impose homogeneous Neumann boundary conditions. With these assumptions, we use FEM for spatial discretization and the backward Euler scheme for time. The FEM basis functions are chosen to be piecewise linear functions with a spatial step size of $h=0.5$, which take a value of 1 at node $i$ and 0 at the other nodes. The time step is set to $0.1$ (see Appendix \ref{sec:FEM} for more details).

\begin{figure}[ht]
    \centering
    \includegraphics[width=\textwidth]{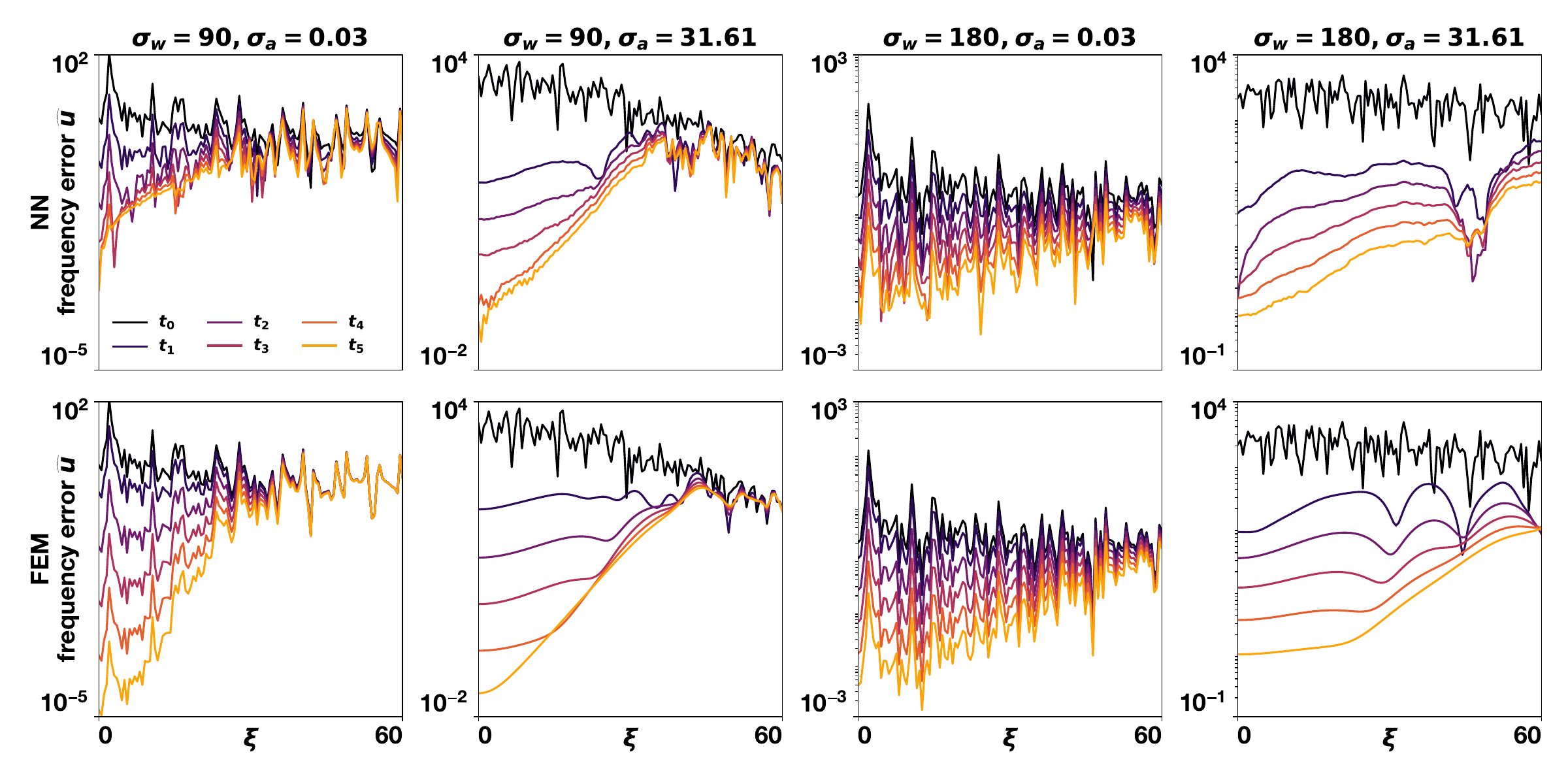}
    \caption{Comparation between FEM simulations of \ref{eq:dynamicsFF} and NN actual dynamics.}
    \label{fig:FEMvsNN}
\end{figure}

In the second experiment, we aim to compare the results obtained by NNs with those obtained by FEM simulations when $\rho_{\bsw}$ follows a normal distribution. To do this, we compute the magnitude of the mean of $\hat{u}$ across the 100 initializations.  
Figure \ref{fig:FEMvsNN} shows that the simulations produced by the FEM (second row) for equation (\ref{eq:dynamicsFF}) exhibit a similar qualitative behavior to the NNs (first row). Additionally, the figure highlights the diffusive effect of the parameter $\sigma_a$. Specifically, the values of $\hat{u}$ tend to become smoother, and the NN is capable of improving the learnability of frequencies at which $\rho_{\bsw}$ is small but positive.

\subsection{Multilayer case}

In addition to our analytical results being based on 2-layer NNs, we aim to demonstrate if these findings also extended to multilayer NNs. The principal assumption that needs to be satisfied is the NTK hypothesis, which posits that the size of the hidden layers tends to infinity. Additionally, the multilayer NN follows the FF model, that is, all the hidden layers have ReLU$(x)=\max(x,0)$ as an activation function, except the first layer which has cosine and sine activation functions.

\begin{figure}[h]
	\centering
		\includegraphics[width=\linewidth]{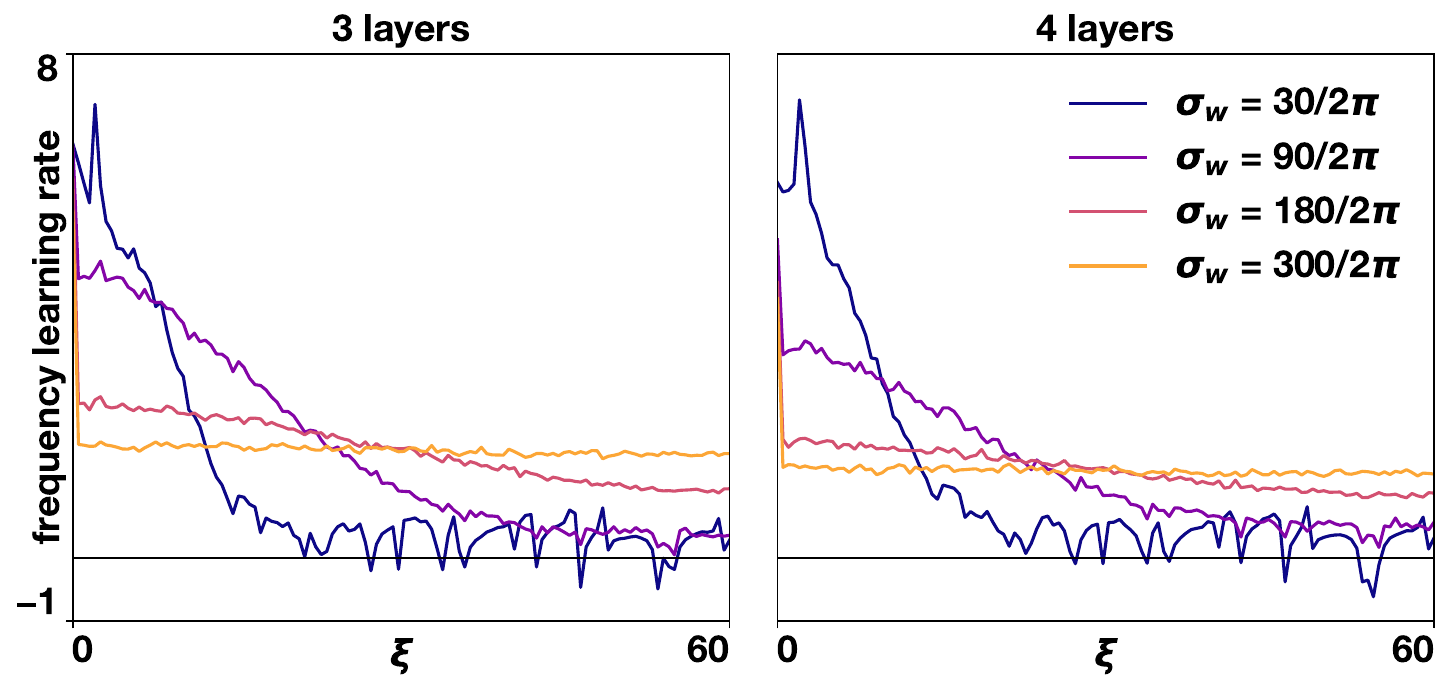}

	\caption{Frequency learning rate $\kappa(\bxi)$ on 3-layer and 4-layer neural networks on NTK regime. The hidden layers have all equal widths of 4000.}\label{fig:Multilayers}
\end{figure}

To study this setting, we analyze the frequency learning rate $\kappa(\bxi)$ given by equation (\ref{eq:learningrate}) for 3-layer and 4-layer NNs across 100 different initializations of Gaussian distribution (See Figure \ref{fig:Multilayers}).

Figure \ref{fig:Multilayers} provides empirical evidence that our results can be extended to multilayer NNs. Indeed, the figure shows behavior similar to the 2-layer case. In both, the 3-layer and 4-layer NNs, the frequency learning rates resembles half of a Gaussian distribution. However, these plots are not identical to those for 2-layer NNs. This difference is particularly marked in their maximum value.

The results of these experiments are consistent with the findings of \cite{yang2021tensor}, which demonstrate that, under the NTK regime, the parameters of the hidden layers tend to remain static. That is, in the infinite-width limit, the hidden layers almost do not change during NN training.

\section{Conclusions and future work}

We rigorously deduced an equation that models the frequency dynamics of a 2-layer NN in the NTK regime, in terms of directly interpretable quantities. Under the additional hypothesis that the NN follows a Fourier Features architecture, the dynamics is explicitly described as the damped heat equation \eqref{eq:dynamicsFF}, where the role of the distributions of the initial parameters is understood in a very precise manner. Namely, the frequency bias of the dynamics is mostly determined by the distribution of the parameter $\rho_{\bsw}$, through the damping term and the diffusion coefficient, while the variance $\sigma_a^2$ of the parameter $a$, only amplifies or reduces the diffusion term in equation \eqref{eq:dynamicsFF}. 

Our theoretical predictions are then validated with numerical experiments. We find qualitative agreement between the dynamics predicted by the FEM solution of our PDE \eqref{eq:dynamicsFF} and the actual FF-NN learning dynamics (figure \ref{fig:FEMvsNN}). Also, we very concretely observe the frequency bias behaviour predicted by equation \eqref{eq:dynamicsFF} in experimental FF-NN frequency learning rates, for different cases of $\rho_{\bsw}$ (figure \ref{fig:learningrate}). Additionally, the theoretical framework developed here for understanding the dynamics of 2-layer FF-NN, seem to extend to somewhat deeper NN (figure \ref{fig:Multilayers}).

By understanding how the distributions $\rho_{\bsw}$ and $\rho_a$ affect the behaviour of equation \eqref{eq:dynamicsFF}, it is possible to predict, and thus eliminate or/and control, frequency bias in the learning dynamics of a FF neural network in the NTK regime. 
If a target data-set follows a known (or estimated) frequency distribution, a promising direction for future work is to apply our model in order to design specific initial distributions that achieve faster convergence to high accuracy approximation of target functions in specific classes.

Finally, in this work the training of the NN is achieved using gradient descent. Using a different optimization scheme might produce a different learning dynamics, and an interesting continuation of our research is to extend the present analysis to other training mechanisms and develop specific optimization strategies that overcome the issue of spectral bias in NN training.

\clearpage
\bibliographystyle{abbrvnat}
\bibliography{references}

\newpage
\appendix
\section{Derivation of spectral bias equation for \(\widehat{u}\)}

The objective of this section is to deduce the frequency dynamics of the two-layer neural network.

In summary, under the assumptions \ref{assu} we arrive at a linearized residual dynamics give by
\begin{equation}\label{eq:dinamicalinealespacioimagen1}
    \frac{du}{dt}(\bsx,\Theta (t))=-\mathbb E_{\theta\sim\rho_a\times\rho_{\bsw}} \left[\int_{\mathbb{R}^{d}}K_\theta\left(\bsx,\bsxp\right)u\left(\bsxp,\Theta (t)\right)d\rho_{\text{data}}(\bsxp)\right],
\end{equation}
where $\rho_{\text{data}}(\bsx):=\sum_{i=1}^{N}\delta_{\bsx_{i}}\left(\bsx\right)$, $u(\bsx,\Theta(t))=f(\bsx,\Theta(t))-\widetilde{f}(\bsx)$ and
\begin{eqnarray}\label{eq:k0}
    K_\theta(\bsx, \bsxp)&:=&\nabla_\theta f^0(\bsx, \theta)\cdot \nabla_\theta f^0(\bsxp,\theta)\nonumber\\
    &=&\partial_a f^0(\bsx,\theta)\partial_a f^0(\bsxp,\theta) + \nabla_{\bsw}f^0(\bsx,\theta) \cdot \nabla_{\bsw} f^0(\bsx,\theta).
\end{eqnarray}
Here, \begin{equation}\label{eq:f0}
    f(x,\Theta)=\frac1{\sqrt m} \sum_{k=1}^m f^0(\bsx,\theta_k) \quad \text{ with } \quad f^0(\bsx,\theta_k):=a_k g(\bsw_k\cdot\bsx).
\end{equation}
for $a_k, \bsw_k$ taken as i.i.d. samples from distributions $\rho_a, \rho_{\bsw}$.

The following computations are based on the theory of tempered distributions. Recall that, over $\mathbb R^d$, Schwartz functions are denoted by $\mathcal S(\mathbb R^d)$ and tempered distributions form the dual space, denoted $\mathcal S'(\mathbb R^d)$. As customary, we say that a relation holds in the sense of distributions over a space $\mathbb R^d$, provided it holds in duality with Schwartz from $\mathcal S(\mathbb R^d)$.

\begin{definition}\label{def:delta-like}
Given a nonzero vector $\bsv \in \mathbb{R}^d$, we define the delta-like distribution $\delta_{\bsv}\in\mathcal S'(\mathbb R^d)$, concentrated on the line spanned by $\bsv$, by requiring
\begin{equation}   
    \left\langle\delta_{\bsv},\phi\right\rangle = \int_\mathbb{R} \phi(y\bsv) dy,\quad\text{for all }\quad\phi \in \mathcal{S}(\mathbb R^d).
\end{equation}
\end{definition}

The Fourier transforms of functions appearing in \eqref{eq:k0} can be computed using the next lemma.
\begin{lemma}\label{lem:fouriercalcs1}
For any $g \in \mathcal{S}'(\R)$ and any $\bsw \in \R^{d}\setminus\{0\}$, the following hold:
\begin{enumerate}
\item[(i)] The Fourier transform of $f_{1}(\bsx)=g(\bsw \cdot \bsx)$ is the distribution
\begin{equation}
\widehat{f_1}(\bxi) = \widehat{g}\left(\frac{\bsw}{\left\Vert \bsw\right\Vert ^{2}} \cdot\bxi\right)\delta_{\bsw}(\bxi).
\end{equation}
\item[(ii)] The Fourier transform of $f_{2}(\bsx)=\bsx g'(\bsw \cdot \bsx)$ is the distribution
\begin{equation}
\widehat{f_2}(\bxi) = -\frac{ia}{2\pi} \nabla_{\bxi} \left[\widehat{g'}\left(\frac{\bsw}{\left\Vert \bsw\right\Vert ^{2}} \cdot\bxi\right)\delta_{\bsw}(\bxi)\right].
\end{equation}
\end{enumerate}
\end{lemma}

\begin{proof}
See \cite{ramasinghe2022frequency} or \cite{luo2022exact}
\end{proof}

\begin{lemma}
The dynamics \eqref{eq:dinamicalinealespacioimagen1} has the following expression in the frequency domain, where $u_{\rho_{\text{data}}}(\bsx,\Theta):=u(\bsx,\Theta) \rho_{\text{data}}(\bsx)$:
\begin{equation}
    \forall \psi\in \mathcal S(\mathbb R^d),\quad\left\langle \partial_t\widehat{u},\psi\right\rangle =      
    \left\langle \mathcal{D}[\widehat{u}_{\rho_{\text{data}}}], \psi \right\rangle,
\end{equation}
where 
\begin{equation}
\mathcal{D}[\widehat{u}_{\rho_{\text{data}}}](\bxi):=-\mathbb{E}_{\theta\sim\rho_a\times\rho_{\bsw}}\int_{\mathbb{R}^{d}}\widehat{K}_\theta\left(\bxi,\bxip\right)\widehat{u}_{\rho_{\text{data}}}\left(\bxip,\Theta\right)d\bxip
= -\mathbb{E}_{\theta} \left\langle \widehat{K}_\theta(\bxi, \cdot),  \overline{\widehat{u}_{\rho_{\text{data}}}}(\cdot) \right\rangle
\end{equation}
and, where, in the sense of distributions over $\mathbb R^d\times\mathbb R^d$, there holds
\begin{alignat}{1}
\widehat{K}_\theta\left(\bxi,\bxip\right):= & \overline{\mathcal{F}_{\bsxp\rightarrow\bxi^{\prime}}\left[\nabla_\theta f^0(\bsxp,\theta)\right]}(\bxi^{\prime})\cdot\mathcal{F}_{\bsx\rightarrow\bxi}\left[\nabla_\theta f^0(\bsx,\theta)\right](\bxi) \nonumber \\
= & \overline{\mathcal{F}_{\bsxp\rightarrow\bxi^{\prime}}\left[\begin{array}{c}
\nabla_{a}f^0(\bsxp,\theta) \nonumber\\
\nabla_{\bsw}f^0(\bsxp,\theta)
\end{array}\right]}(\bxi^{\prime})\cdot\mathcal{F}_{\bsx\rightarrow\bxi}\left[\begin{array}{c}
\nabla_{a}f^0(\bsx,\theta)\\
\nabla_{\bsw}f^0(\bsx,\theta)
\end{array}\right](\bxi) \nonumber\\
:= & \widehat{K}_{a}\left(\bxi,\bxip\right)+\widehat{K}_{\bsw}\left(\bxi,\bxip\right) \label{eq:ktheta1}
\end{alignat}
\end{lemma}

\begin{proof}
See \cite{luo2022exact}.
\end{proof}

\begin{theorem}[cf. Thm. \ref{thm:evol}]\label{thm:evol_app}
Under the assumptions \ref{assu}, the dynamics (\ref{eq:dinamicalinealespacioimagen1}) can be expressed in frequency space, in the sense of distributions over $\mathbb R^d$, as:
\begin{alignat}{1}
\left\langle \frac{d}{dt}\widehat{u},\psi\right\rangle 
= & -\frac{\sigma_{a}^{2}}{4\pi^{2}} \int_{\mathbb{R}^{d}}   \overline{\nabla\psi(\bxi)} \cdot \int_{\mathbb{R}^{2}} \nabla{\widehat{u}_{\rho_{\text{data}}}\left(\frac{y'}{y}\bxi\right)}    \frac{\overline{\widehat{g'}\left(y'\right)} \widehat{g'}\left(y\right)}{\lvert y \rvert^d}dy'dy\ \rho_{\bsw}\left(\frac{1}{y}\bxi\right) d\bxi \label{eq:dynamicsfinal1}\\
 & -\int_{\mathbb{R}^{d}}   \overline{\psi(\bxi)} \int_{\mathbb{R}^{2}} {\widehat{u}_{\rho_{\text{data}}}\left(\frac{y'}{y}\bxi\right)}   \frac{\overline{\widehat{g}\left(y'\right)} \widehat{g}\left(y\right)}{\lvert y \rvert^d}dy'dy\ \rho_{\bsw}\left(\frac{1}{y}\bxi\right)d\bxi . \nonumber
\end{alignat}

\end{theorem}

\begin{proof}
Extending the computations from \cite{luo2022exact}, we start by writing for $\phi,\psi\in\mathcal S(\mathbb R^d)$ 
\begin{equation}
\left\langle \widehat{K}_{\bsw},\overline{\phi} \otimes \psi\right\rangle :=
\left\langle \left\langle\widehat{K}_\theta,\overline{\phi}\right\rangle, \psi\right\rangle=\int_{\mathbb{R}^{2d}}\widehat{K}_\theta\left(\bxi,\bxip\right)\overline{\psi(\bxi)}{\phi(\bxip)}d\bxi d\bxip.
\end{equation}

As $\widehat{K}_\theta=\widehat{K}_{\bsw}+\widehat{K}_{a}$ (cf. \eqref{eq:ktheta1}), we treat the two terms separately. For $\widehat{K}_{\bsw}$ we have, using Lemma \ref{lem:fouriercalcs1}, 
\begin{alignat}{1}
&\left\langle \widehat{K}_{\bsw},\overline{\phi} \otimes \psi\right\rangle \\
& =\int_{\mathbb{R}^{2d}} \overline{\mathcal{F}_{\bsxp\rightarrow\bxi^{\prime}} \left[\nabla_{\bsw}f^0(\bsxp,\theta)\right]}(\bxi^{\prime}) \cdot \mathcal{F}_{\bsx\rightarrow\bxi}\left[\nabla_{\bsw}f^0(\bsx,\theta)\right](\bxi) \overline{\psi(\bxi)}{\phi(\bxip)}d\bxip 
 d\bxi \nonumber\\
 & =\int_{\mathbb{R}^{2d}} \overline{\mathcal{F}_{\bsxp\rightarrow\bxi^{\prime}} \left[a\ \bsx^{\prime} g'(\bsw \cdot \bsx^{\prime})\right]}(\bxi^{\prime}) \cdot \mathcal{F}_{\bsx\rightarrow\bxi}\left[a\ \bsx g'(\bsw \cdot \bsx)\right](\bxi) \overline{\psi(\bxi)}{\phi(\bxip)}d\bxip 
 d\bxi \nonumber \\
& =\int_{\mathbb{R}^{2d}} \frac{ia}{2\pi} \nabla_{\bxip}\left[\overline{\widehat{g'}\left(\frac{\bsw}{\left\Vert \bsw\right\Vert ^{2}} \cdot \bxip\right)\delta_{\bsw}(\bxip)}\right] \cdot-\frac{ia}{2\pi} \nabla_{\bxi} \left[\widehat{g'}\left(\frac{\bsw}{\left\Vert \bsw\right\Vert ^{2}} \cdot\bxi\right)\delta_{\bsw}(\bxi)\right] \overline{\psi(\bxi)}{\phi(\bxip)} d\bxip d\bxi \nonumber\\
 & =\frac{a^2}{4\pi^{2}} \int_{\mathbb{R}^{d}} \nabla_{\bxip} \left[\overline{\widehat{g'}\left(\frac{\bsw}{\left\Vert \bsw\right\Vert ^{2}} \cdot\bxip\right)} \delta_{\bsw}(\bxip)\right] {\phi(\bxip)}d\bxip \cdot\int_{\mathbb{R}^{d}} \nabla_{\bxi}\left[\widehat{g'} \left(\frac{\bsw}{\left\Vert \bsw\right\Vert ^{2}} \cdot\bxi\right)\delta_{\bsw}(\bxi)\right] \overline{\psi(\bxi)}d\bxi \nonumber\\
 & =\frac{a^2}{4\pi^{2}} \int_{\mathbb{R}^{d}}\delta_{\bsw}(\bxip) \overline{\widehat{g'}\left(\frac{\bsw}{\left\Vert \bsw\right\Vert ^{2}}\cdot\bxip\right)} \nabla_{\bxip}{\phi(\bxip)}d\bxip  \cdot \int_{\mathbb{R}^{d}}\delta_{\bsw}(\bxi) \widehat{g'}\left(\frac{\bsw}{\left\Vert \bsw\right\Vert ^{2}} \cdot\bxi\right)\nabla_{\bxi}\overline{\psi(\bxi)} d\bxi.\nonumber
\end{alignat}
Hence, 
\begin{alignat}{1}
\left\langle \widehat{K}_{\bsw},\overline{\phi}\otimes\psi\right\rangle  & =\frac{a^2}{4\pi^{2}}\int_{\mathbb{R}}\overline{\widehat{g'}\left(\frac{\bsw}{\left\Vert \bsw\right\Vert ^{2}}\cdot y'\bsw \right)}\nabla{\phi(y'\bsw )} dy'\cdot\int_{\mathbb{R}}\widehat{g'}\left(\frac{\bsw}{\left\Vert \bsw\right\Vert ^{2}}\cdot y\bsw \right)\nabla\overline{\psi(y\bsw)}dy \nonumber\\
 & =\frac{a^2}{4\pi^{2}}\int_{\mathbb{R}^{2}}\nabla\overline{\psi(y\bsw)}\cdot\nabla{\phi(y'\bsw )} \overline{\widehat{g'}\left(y'\right)}\widehat{g'}\left(y\right)dydy'. 
\end{alignat}
Next, taking the expectation over $\theta\sim \rho_a\times\rho_{\bsw}$, we get: 
\begin{alignat}{1}   
\mathbb{E}_\theta\left\langle \widehat{K}_{\bsw},\overline{\phi}\otimes\psi\right\rangle & 
=\frac{\sigma_{a}^{2}}{4\pi^{2}}\int_{\mathbb R^d}\int_{\mathbb{R}^2} \nabla \overline{\psi(y\bsw)} \cdot \nabla{\phi(y'\bsw )} \overline{\widehat{g'}\left(y'\right)} \widehat{g'}\left(y\right)dy'dy\rho_{\bsw}(\bsw)d\bsw. 
\end{alignat}
Next, we use the following change of variable: 
\begin{equation}
\bseta =y\bsw=y\left(w_{1},\ldots,w_{d}\right)^{T}, \quad \left|\mathrm{det}\left(\frac{\partial(w_1,\dots,w_d)}{\partial(\eta_1,\dots,\eta_d)}\right)\right|=\frac{1}{|y|^d},\label{eq:change-variable-radially1} \nonumber 
\end{equation}
obtaining
\begin{alignat}{1}
  \mathbb{E}_\theta\left\langle \widehat{K}_{\bsw},\overline{\phi}\otimes\psi\right\rangle 
= & \frac{\sigma_{a}^{2}}{4\pi^{2}} \int_{\mathbb{R}^{d}} \int_{\mathbb{R}^{2}}  \nabla\overline{\psi(\bseta)} \cdot \nabla{\phi\left(\frac{y'}{y}\bseta\right)} \overline{\widehat{g'}\left(y'\right)} \widehat{g'}\left(y\right) \rho_{\bsw}\left(\frac{1}{y}\bseta\right)\frac{1}{\lvert y \rvert^d}dy'dyd\bseta \\
= & \frac{\sigma_{a}^{2}}{4\pi^{2}} \int_{\mathbb{R}^2} \int_{\mathbb{R}^d} \nabla\overline{\psi(\bseta)} \cdot \left(\nabla{\phi\left(\frac{y'}{y}\bseta\right)} \rho_{\bsw}\left(\frac{1}{y}\bseta\right) \right) d\bseta \overline{\widehat{g'}\left(y'\right)} \widehat{g'}\left(y\right) \frac{1}{\lvert y \rvert^d}dy'dy. \nonumber
\end{alignat}
Then, applying the divergence theorem, we obtain  
\begin{alignat}{1}\label{eq:expectationW}
  \mathbb{E}_\theta\left\langle \widehat{K}_{\bsw},\overline{\phi}\otimes\psi\right\rangle 
= & -\frac{\sigma_{a}^{2}}{4\pi^{2}} \int_{\mathbb{R}^{d}}   \overline{\psi(\bseta)}\int_{\mathbb{R}^{2}} \text{div}\left(\nabla{\phi\left(\frac{y'}{y}\bseta\right)} \rho_{\bsw}\left(\frac{1}{y}\bseta\right) \right)  \frac{\overline{\widehat{g'}\left(y'\right)} \widehat{g'}\left(y\right)}{\lvert y \rvert^d}dy'dy d\bseta.
\end{alignat}

Similarly, for $K_a$ we calculate
\begin{alignat}{1}
\left\langle \widehat{K}_{a},\overline{\phi} \otimes \psi\right\rangle  & =\int_{\mathbb{R}^{2d}} \overline{\mathcal{F}_{\bsxp\rightarrow\bxi^{\prime}} \left[\nabla_{a}f^0(\bsxp,\theta)\right]}(\bxi^{\prime}) \cdot \mathcal{F}_{\bsx\rightarrow\bxi}\left[\nabla_{a}f^0(\bsx,\theta)\right](\bxi) \overline{\psi(\bxi)}{\phi(\bxip)} d\bxip d\bxi\\
& =\int_{\mathbb{R}^{2d}} \overline{\mathcal{F}_{\bsxp\rightarrow\bxi^{\prime}} \left[g(\bsw \cdot \bsxp)\right]}(\bxi^{\prime}) \cdot \mathcal{F}_{\bsx\rightarrow\bxi}\left[g(\bsw \cdot \bsx)\right](\bxi) \overline{\psi(\bxi)}{\phi(\bxip)}d\bxip d\bxi \nonumber\\
 & = \int_{\mathbb{R}^{2d}}\overline{\widehat{g}\left(\frac{\bsw}{\left\Vert \bsw\right\Vert ^{2}}\cdot\bxip\right)}\delta_{\bsw}(\bxip)\widehat{g}\left(\frac{\bsw}{\left\Vert \bsw\right\Vert ^{2}}\cdot\bxi\right)\delta_{\bsw}(\bxi) {\phi(\bxi)}\overline{\psi(\bxip)}d\bxip d\bxi \nonumber\\
 & = \int_{\mathbb{R}^{2}} \overline{\psi(y'\bsw )} {\phi(y\bsw)} \overline{\widehat{g}\left(y'\right)}\widehat{g}\left(y\right)dy'dy. \nonumber
\end{alignat}
Applying the expectation with respect to $\theta$ and using the same change of variable as before, we get 
\begin{alignat}{1}\label{eq:Aexpectation}
  \mathbb{E}_\theta\left\langle \widehat{K}_{a},\overline{\phi}\otimes\psi\right\rangle 
= &  \int_{\mathbb{R}^2}  \int_{\mathbb{R}^d} \overline{\psi(\bseta)}  {\phi\left(\frac{y'}{y}\bseta\right)} \rho_{\bsw}\left(\frac{1}{y}\bseta\right)  d\bseta \ \overline{\widehat{g}\left(y'\right)} \widehat{g}\left(y\right) \frac{1}{\lvert y \rvert^d}dy'dy. 
\end{alignat}

Combining equations (\ref{eq:expectationW}) and (\ref{eq:Aexpectation}) and taking $\phi=\widehat{u}_{\rho_{\text{data}}}$, we obtain
\begin{alignat}{1}
  \left\langle \frac{d}{dt}\widehat{u},\psi\right\rangle  = & -\mathbb{E}_\theta\left\langle \widehat{K}_\theta,\overline{\widehat{u}_{\rho_{\text{data}}}}\otimes\psi\right\rangle 
 =  -\mathbb{E}_\theta\left\langle  \widehat{K}_{\bsw} + \widehat{K}_{a},\overline{\widehat{u}_{\rho_{\text{data}}}}\otimes\psi\right\rangle \nonumber\\
= & \frac{\sigma_{a}^{2}}{4\pi^{2}} \int_{\mathbb{R}^d}   \overline{\psi(\bseta)}\int_{\mathbb{R}^2} \text{div}\left(\nabla{\widehat{u}_{\rho_{\text{data}}}\left(\frac{y'}{y}\bseta\right)} \rho_{\bsw}\left(\frac{1}{y}\bseta\right) \right) \frac{\overline{\widehat{g'}\left(y'\right)} \widehat{g'}\left(y\right)}{\lvert y \rvert^d}dy'dy d\bseta \nonumber\\
 & -\int_{\mathbb{R}^d}   \overline{\psi(\bseta)} \int_{\mathbb{R}^2} {\widehat{u}_{\rho_{\text{data}}}\left(\frac{y'}{y}\bseta\right)} \rho_{\bsw}\left(\frac{1}{y}\bseta\right) \frac{\overline{\widehat{g}\left(y'\right)} \widehat{g}\left(y\right)}{\lvert y \rvert^d}dy'dy d\bseta, \nonumber
\end{alignat}

which concludes the proof.
\end{proof}

\section{Derivation of spectral bias for the Fourier Features Model}

In the case of a Fourier Features model, recall that our neural network is defined by 
\begin{equation}\label{FFmodel1}
f(\bsx,\theta(t)) =\frac{1}{\sqrt{2m}}\sum_{k=1}^{m}a_{k}(t)g_1\left(\bsw_{k}(t)\cdot \bsx\right)+\frac{1}{\sqrt{2m}}\sum_{k=1}^{m}b_{k}(t)g_2\left(\bsw_{k}(t)\cdot \bsx\right),
\end{equation}
where $\theta(t)=\{\bsa (t), \bsb (t), W(t)\}$, with $\bsa \in \R^{m}$, $\bsb \in \R^{m}$ and $W(t)=(\bsw_1(t), \dots, \bsw_m(t))^T \in \R^{d}\times \R^{m}$, are the trainable parameters having initial distributions $\rho_{a}$, $\rho_{b}$ and $\rho_{\bsw}$,  respectively. In this case, the NTK calculation using Lemma \ref{lem:fouriercalcs1} gives 
\begin{alignat}{1}
K(\bsx,\bsxp)= & \mathbb{E}_\theta\nabla_{\theta}f^0(\bsx,\theta)\cdot\nabla_{\theta}f^0(\bsxp,\theta)\\
= & \mathbb{E}_\theta\left[a^{2}g_{1}'\left(\bsw\cdot \bsx\right)g_{1}'\left(\bsw\cdot \bsxp\right)\bsx\cdot \bsxp+b^{2}g_{2}'\left(\bsw\cdot \bsx\right)g_{2}'\left(\bsw\cdot \bsxp\right)\bsx\cdot \bsxp\right. \nonumber\\
 & +abg_{1}'\left(\bsw\cdot \bsx\right)g_{2}'\left(\bsw\cdot \bsxp\right)\bsx\cdot \bsxp+abg_{2}'\left(\bsw\cdot \bsx\right)g_{1}'\left(\bsw\cdot \bsxp\right)\bsx\cdot \bsxp \nonumber\\
 & \left.+g_{1}\left(\bsw\cdot \bsx\right)g_{1}\left(\bsw\cdot \bsxp\right)+g_{2}\left(\bsw\cdot \bsx\right)g_{2}\left(\bsw\cdot \bsxp\right)\right] \nonumber \\
 = & \mathbb{E}_\theta\left[a^{2}g_{1}'\left(\bsw\cdot \bsx\right)g_{1}'\left(\bsw\cdot \bsxp\right)\bsx\cdot \bsxp\right]+ \mathbb{E}_\theta \left[b^{2}g_{2}'\left(\bsw\cdot \bsx\right)g_{2}'\left(\bsw\cdot \bsxp\right)\bsx\cdot \bsxp\right] \nonumber\\
 & +\mathbb{E}_\theta\left[g_{1}\left(\bsw\cdot \bsx\right)g_{1}\left(\bsw\cdot \bsxp\right)\right]+ \mathbb{E}_\theta\left[g_{2}\left(\bsw\cdot \bsx\right)g_{2}\left(\bsw\cdot \bsxp\right)\right] \nonumber
\end{alignat}

\begin{corollary}[Cf. Cor. \ref{cor:PDE}]\label{cor:PDE_app}
Under the assumption \ref{assu}. If the NN is given by equation (\ref{FFmodel}) where $g_1(x)=\cos(2\pi x)$, $g_2(x)=\sin(2\pi x)$, $\mathbb{E}[a^2]=\mathbb{E}[b^2]=\sigma_a^2$ and $\rho_{\bsw}$ is an 
even function, then the dynamics (\ref{eq:dynamicsfinal1}) in the frequency space is given by
\begin{equation}
\frac{d}{dt}\widehat{u}(\bxi,\Theta(t)) 
= \sigma_{a}^{2} \mathrm{div}\left(\rho_{\bsw}(\bxi)\nabla{\widehat{u}_{\rho_{\text{data}}}\left(\bxi, \Theta(t)\right)}  \right) - \rho_{\bsw}(\bxi){\widehat{u}_{\rho_{\text{data}}}\left(\bxi,\Theta(t)\right)}  
\label{eq:dynamicsFF1}
\end{equation}
\end{corollary}

\begin{proof}
It directly follows that

\begin{alignat}{2}
    \left\langle \frac{d}{dt}\widehat{u},\psi\right\rangle= & -\mathbb{E}_\theta\left\langle \widehat{K}_{a,\bsw}+\widehat{K}_{b,\bsw}+\widehat{K}_{a}+\widehat{K}_{b},\overline{\widehat{u}_{\rho_{\text{data}}}}\otimes\psi\right\rangle\\ 
    = & -\frac{\sigma_{a}^{2}}{4\pi^{2}}\int_{\mathbb{R}^{d}} \overline{\nabla\psi(\bxi)} \cdot \int_{\mathbb{R}^{2}} \nabla{\widehat{u}_{\rho_{\text{data}}}\left(\frac{y'}{y}\bxi\right)} \rho_{\bsw}\left(\frac{1}{y}\bxi\right)   \frac{\overline{\widehat{g'_1}\left(y'\right)} \widehat{g'_1}\left(y\right)}{\lvert y \rvert^d}dy'dy d\bxi \nonumber\\
    & -\frac{\sigma_{b}^{2}}{4\pi^{2}}\int_{\mathbb{R}^{d}} \nabla\overline{\psi(\bxi)} \cdot \int_{\mathbb{R}^{2}} \nabla{\widehat{u}_{\rho_{\text{data}}}\left(\frac{y'}{y}\bxi\right)} \rho_{\bsw}\left(\frac{1}{y}\bxi\right)   \frac{\overline{\widehat{g'_2}\left(y'\right)} \widehat{g'_2}\left(y\right)}{\lvert y \rvert^d}dy'dy d\bxi \nonumber\\
    & -\int_{\mathbb{R}^{d}}\overline{\psi(\bxi)} \int_{\mathbb{R}^{2}} {\widehat{u}_{\rho_{\text{data}}}\left(\frac{y'}{y}\bxi\right)} \rho_{\bsw}\left(\frac{1}{y}\bxi\right)   \frac{\overline{\widehat{g_1}\left(y'\right)} \widehat{g_1}\left(y\right)}{\lvert y \rvert^d}dy'dy d\bxi \nonumber \\
    & -\int_{\mathbb{R}^{d}}\overline{\psi(\bxi)} \int_{\mathbb{R}^{2}} {\widehat{u}_{\rho_{\text{data}}}\left(\frac{y'}{y}\bxi\right)} \rho_{\bsw}\left(\frac{1}{y}\bxi\right)   \frac{\overline{\widehat{g_2}\left(y'\right)} \widehat{g_2}\left(y\right)}{\lvert y \rvert^d}dy'dy d\bxi \nonumber
\end{alignat}

As
\begin{alignat}{2}
    \widehat{g}_1(z) = & \frac{1}{2}\left(\delta(z-1)+\delta(z+1)\right)
    \quad \text{and} \quad & \widehat{g'_{1}}(z)=-\frac{\pi}{i}\left(\delta(z-1)-\delta(z+1)\right) \\
    \widehat{g_2}(z) = & \frac{1}{2i}\left(\delta(z-1)-\delta(z+1)\right)
    \quad \text{and} \quad  & \widehat{g_{2}^{'}}(z)=\pi\left(\delta(z-1)+\delta(z+1)\right), \nonumber
\end{alignat}
then
 \begin{alignat}{1}
& \left\langle \frac{d}{dt}\widehat{u},\psi\right\rangle \\ = & -\int_{\mathbb{R}^{d}}\left[\frac{\sigma_{a}^{2}}{4} \nabla\overline{\psi(\bxi)}\cdot \left(2\nabla{\widehat{u}_{\rho_{\text{data}}} \left(\bxi\right)}+2\nabla{\widehat{u}_{\rho_{\text{data}}}\left(-\bxi\right)}\right)+\frac{\sigma_{b}^{2}}{4}\nabla\overline{\psi(\bxi)} \cdot\left(2\nabla{\widehat{u}_{\rho_{\text{data}}}\left(\bxi\right)}-2\nabla{\widehat{u}_{\rho_{\text{data}}}\left(-\bxi\right)}\right)\right. \nonumber \\
 & + \left.\frac{1}{4}\overline{\psi(\bxi)} \left(2{\widehat{u}_{\rho_{\text{data}}}\left(\bxi\right)}-2{\widehat{u}_{\rho_{\text{data}}}\left(-\bxi\right)}\right)
  + \frac{1}{4}\overline{\psi(\bxi)}\left(2{\widehat{u}_{\rho_{\text{data}}}\left(\bxi\right)} +2{\widehat{u}_{\rho_{\text{data}}}\left(-\bxi\right)}\right)\right]\rho_{\bsw}( \bxi)d\bxi, \nonumber
\end{alignat}
therefore 
\begin{equation}
\left\langle \frac{d}{dt}\widehat{u},\psi\right\rangle  =- \int_{\mathbb{R}^{d}}\left(\sigma_{a}^{2} \nabla\overline{\psi(\bxi)}\cdot\nabla {\widehat{u}_{\rho_{\text{data}}}\left(\bxi\right)}+ \overline{\psi(\bxi)}{\widehat{u}_{\rho_{\text{data}}} \left(\bxi\right)}\right)\rho_{\bsw}( \bxi)d\bxi
\end{equation}
and
\begin{equation}
\frac{d}{dt}\widehat{u}(\bxi,\Theta(t)) 
= \sigma_{a}^{2} \mathrm{div}\left(\rho_{\bsw}(\bxi)\nabla{\widehat{u}_{\rho_{\text{data}}}\left(\bxi, \Theta(t)\right)}  \right) - \rho_{\bsw}(\bxi) {\widehat{u}_{\rho_{\text{data}}}\left(\bxi,\Theta(t)\right)}  
\end{equation}
\end{proof}

\begin{remark}
If $g_1(x)=\cos(cx)$ and $g_2(x)=\sin(c x)$, then 
\begin{equation}\label{eq:dimensionaldependence}
\frac{d}{dt}\widehat{u}(\bxi,\Theta(t))  = \left(\frac{2\pi}{c}\right)^d \left[\frac{c^2 \sigma_{a}^{2}}{4\pi^2} \mathrm{div}\left(\rho_{\bsw}\left( \frac{2\pi}{c}\bxi\right) \nabla{ \widehat{u}_{\rho_{\text{data}}}\left(\bxi, \Theta(t)\right)}  \right)-  \rho_{\bsw}\left( \frac{2\pi}{c}\bxi\right){\widehat{u}_{\rho_{\text{data}}} \left(\bxi\right)}\right].
\end{equation}
\end{remark}

\begin{remark}
If $\rho_{\bsw}$ is not even, then the dynamics is given as follows, 
\begin{alignat}{1}
    \frac{d}{dt}\widehat{u}(\bxi,\Theta(t)) 
 = & \sigma_{a}^{2} \mathrm{div}\left[\frac{\left(\rho_{\bsw}(\bxi)  + \rho_{\bsw}(-\bxi)\right)}{2}
\nabla{\widehat{u}_{\rho_{\text{data}}}\left(\bxi, \Theta(t)\right)}  \right] \label{eq:noevendistribution} \\
 &-\frac{\left(\rho_{\bsw}(\bxi) + \rho_{\bsw}(-\bxi)\right)}{2} {\widehat{u}_{\rho_{\text{data}}}\left(\bxi,\Theta(t)\right)} . \nonumber
\end{alignat}
\end{remark}

\section{Numerical solution of the damped heat equation \eqref{eq:dynamicsFF}.}\label{sec:FEM}

In order to validate our findings we experimentally compared the behavior of the exact learning dynamics of a FF neural network, with the theoretical prediction given by equation (\ref{eq:dynamicsFF}). To solve equation \eqref{eq:dynamicsFF} we used a standard approach with FEM and Backward Euler scheme, as mentioned below.

\subsection{Finite element approximation}\label{FEMapprox}

Let $H^{1}(\Omega)$ be the standard Sobolev space, for
$\Omega\subset\mathbb{R}$, with norm
\begin{equation}
\|\phi\|_{H^{1}}:=\left(\int_\Omega |\phi(x)|^{2}+|\partial_{x}\phi(x)|^{2} dx\right)^{1/2},
\end{equation}
with derivatives considered in the weak sense (distributional). 

If we let $\phi\in H^{1}(\Omega)$ be a test function, multiplying equation  by $\phi$ and integrating over $\Omega$, using integration by parts and the fact that
$\partial_{\xi}u(\xi,t)=0$ on the boundary, we obtain the variational form of \eqref{eq:dynamicsFF}:  
\begin{align}
\int_{\Omega}\partial_{t}\hat{u}(\xi,t) \overline{\phi(\xi)}\,\mathrm{d}\xi & =  -\sigma_a^{2}\int_{\Omega}\rho_{\bsw}(\xi)\partial_{\xi} {\hat{u}(\xi,t)} \partial_{\xi} \overline{\phi(\xi)}\,\mathrm{d}\xi-\int_{\Omega}\rho_{\bsw}(\xi){\hat{u}(\xi,t)} \overline{\phi(\xi)}\mathrm{d}\xi. \quad t\geq 0,
\label{eq:formulacionvariacional}
\end{align}
with an initial condition prescribing the value of $\hat u(\xi,0)$. Given $T>0$, the problem is to find $u\in C^1([0,T];H^{1}(\Omega))$, such that \eqref{eq:formulacionvariacional} holds for all $\phi\in H^{1}(\Omega)$.

Let $\tau_{h}$ be a partition of $\Omega$ with diameter $h$, and define the corresponding standard finite element space $U_h  \in H^{1}$ of dimension $N$. The numerical scheme goal is to find an approximated solution $u(\xi,t)$ of equation \eqref{eq:formulacionvariacional} of the form
\begin{equation}
u(\xi,t)\approx\sum_{1\leq j\leq N}c_{j}(t)\varphi_{j}(\xi),\label{eq:solaproximada}
\end{equation}
where $\left\{ \varphi_{i}(x)\right\} _{i=1}^{N}$ is a real-valued basis of $U_h$ and $c_i(t) \in \C$ for all $t\in[0,T]$. Substituting each $\overline\phi=\overline\varphi_j=\varphi_j$ in equation \eqref{eq:formulacionvariacional} we obtain a system of equations given by  
\begin{equation}\label{eq:coeffsfem}
\sum_{j=1}^{N}c_{j}^{\prime}(t)\int_{\Omega}\varphi_{j}\varphi_{i}\,\mathrm{d}\xi=-\sigma_a^{2}\sum_{j=1}^{N}c_{j}(t)\int_{\Omega}\rho_{\bsw}(\xi)\nabla\varphi_{j}\nabla\varphi_{i}\,\mathrm{d}\xi-\sum_{j=1}^{N}c_{j}(t)\int_{\Omega}\rho_{\bsw}(\xi)\varphi_{j}\varphi_{i}\,\mathrm{d}\xi,
\end{equation}
for $t\in[0,T]$ and with $i=1,...,N$, and where $c_j(t)$ are unknown functions to be determined.

In summary, if we let $c(t)=(c_{1}(t),c_{2}(t),\ldots,c_{N}(t))$ be the vector of unknown time coefficients, the system of equations for $c$ can be rewritten as: 
\begin{equation}
Lc'(t)  =  - \sigma_a^{2}Mc(t)-Nc(t),\label{eq:sismatricial}
\end{equation}
where 
\begin{equation*}
  (L)_{i,j}  = \int_{\Omega}\varphi_{j}\varphi_{i}\,\mathrm{d}\xi, \qquad 
  (M)_{i,j}  =  \int_{\Omega}\rho_{\bsw}\nabla\varphi_{j}\nabla\varphi_{i}\,\mathrm{d}\xi,\quad \text{ and }  \quad 
      (N)_{i,j}  =  \int_{\Omega}\rho_{\bsw} \varphi_{j}\varphi_{i}\,\mathrm{d}\xi.\
\end{equation*}

Finally, we discretize the time variable using a Backward Euler scheme: let $t_0=0 < t_1 < ... < t_{n_\mathrm{t}}=T$ a partition of the time interval with into segments of length $h_t$ and let $c_k\approx c(t_k), k=1,...,{n_\mathrm{t}}$, then for the fully discrete equation, with our choice of $h_t$ and with $h$ as above, is 
\begin{equation}
    (L+ h_t\sigma_a^{2}M + h_tN)c_{k+1}  =  Lc_k
\end{equation}
for $k=0,...,{n_\mathrm{t}}$, where $c_0$ is obtained similarly to \eqref{eq:coeffsfem} from the initial condition $\hat{u}_0=\hat u(\xi,0)$.

For our experiment were chose $T=500$ time step ${h_\mathrm{t}}=0.1$ and $\Omega = (-60, 60)$ with spatial step $h=0.5$. For each index $i$, the basis function $c_i$ for our FEM implementation is piecewise linear, and takes the value $1$ at node $i$ and $0$ at the other nodes.

To compare to NN training, we use different time snapshots in Figure \ref{fig:FEMvsNN}. The NN snapshot is taken every 4000 iterations, whereas the FEM snapshot is taken every 500-time steps.

\section{Other numerical experiments}

\subsection{Robustness of assumption: NTK}

In this subsection we empirically test the robustness of the NTK assumption, more precisely we test how large must be $m$ to ensure that our results remain qualitatively valid.

\begin{figure}[h]\centering{}\includegraphics[width=\textwidth]{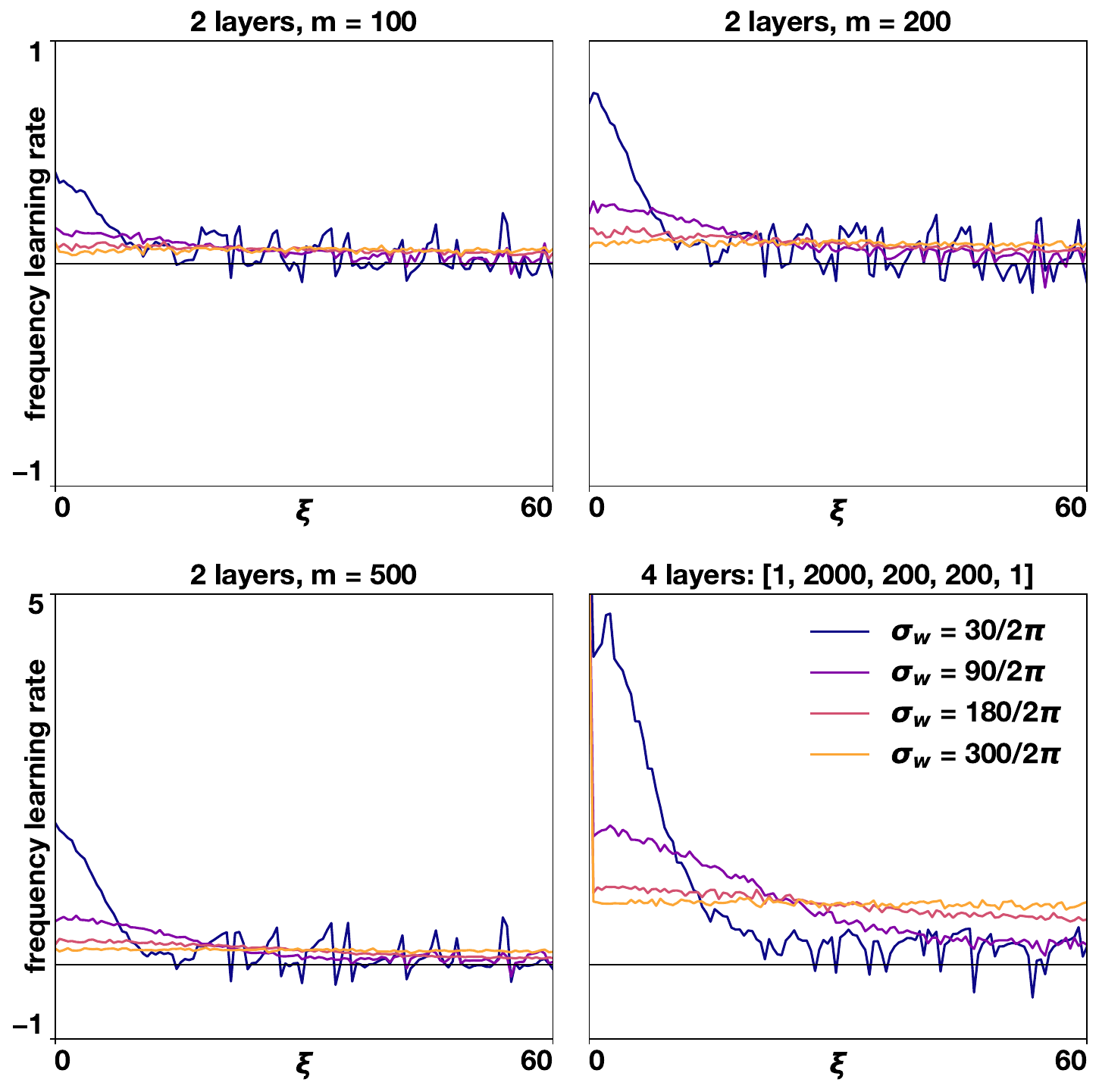}\caption{Robustness analysis for the NTK assumption with different values of $m$ and number of layers.}\label{fig:robustness}
\end{figure}

In Figure \ref{fig:robustness} we present a qualitative analysis of our model for deeper architectures and smaller hidden layer widths $m$. We find that the frequency learning rate $\kappa$ still depends on the initial distribution despite the fact that $m$ has a low value. Furthermore, the hyperparameter $m$ plays an important role in the learning speed. Indeed, note the change in order of magnitudes of frequency learning rates for different values of $m$, compared with figure \ref{fig:learningrate}, where the learning speed reaches orders of $8$ while for small values of $m$ is close to $2$. The same holds for the multilayer case (see the 4-layer of the figure \ref{fig:Multilayers} vs the 4-layer of the figure \ref{fig:robustness}).


\end{document}

%% file: preamble.tex
\usepackage{amsmath}
\usepackage{amssymb}
\usepackage{mathtools}
\usepackage{amsthm}

\theoremstyle{plain}
\newtheorem{theorem}{Theorem}[section]

\newtheorem{lemma}[theorem]{Lemma}
\newtheorem{corollary}[theorem]{Corollary}
\theoremstyle{definition}
\newtheorem{definition}[theorem]{Definition}

\theoremstyle{remark}
\newtheorem{remark}[theorem]{Remark}

\usepackage{graphicx}
\usepackage{tikz}
\usepackage{subfigure}
\usepackage{subcaption}
\usepackage{caption}
\usepackage[export]{adjustbox}
\usepackage{pgfplots}
\pgfplotsset{compat=1.17}
\usepackage[utf8]{inputenc} 
\usepackage[T1]{fontenc}    
\usepackage{hyperref}       
\usepackage{url}            
\usepackage{booktabs}       
\usepackage{amsfonts}       
\usepackage{nicefrac}       
\usepackage{microtype}      
\usepackage{xcolor}         
\usepackage{overpic}

\newcommand{\R}{\mathbb{R}} 
\newcommand{\C}{\mathbb{C}} 

\newcommand{\bxi}{\boldsymbol{\xi}}
\newcommand{\bxip}{\boldsymbol{\xi'}}

\newcommand{\bseta}{\boldsymbol{\eta}}

\newcommand{\bsa}{\boldsymbol{a}}
\newcommand{\bsb}{\boldsymbol{b}}

\newcommand{\bsv}{\boldsymbol{v}}
\newcommand{\bsx}{\boldsymbol{x}}
\newcommand{\bsxp}{\boldsymbol{x'}}
\newcommand{\bsw}{\boldsymbol{w}}

%% file: plots/FFmodel.tex
\begin{figure}[ht]
    \centering                
    
    \begin{tikzpicture}[x=0.75pt,y=0.75pt,yscale=-1,xscale=1]
    
    \draw  [color={rgb, 255:red, 0; green, 0; blue, 0 }  ,draw opacity=1 ] (124.72,121.83) -- (149.09,121.83) -- (149.09,159.83) -- (124.72,159.83) -- cycle ;
    \draw  [color={rgb, 255:red, 0; green, 0; blue, 0 }  ,draw opacity=1 ] (200,79.07) -- (282,79.07) -- (282,201.07) -- (200,201.07) -- cycle ;
    \draw    (345,139.6) ;
    \draw  [color={rgb, 255:red, 189; green, 16; blue, 224 }  ,draw opacity=1 ][fill={rgb, 255:red, 235; green, 220; blue, 239 }  ,fill opacity=1 ] (200,79.07) -- (200,201.07) -- (150.09,159.83) -- (149.09,121.83) -- cycle ;
    \draw   (341.06,140.33) -- (282,201.07) -- (282,79.07) -- cycle ;
    \draw  [dash pattern={on 4.5pt off 4.5pt}]  (149.59,140.83) -- (282,140.45) ;
    \draw  [color={rgb, 255:red, 0; green, 0; blue, 0 }  ,draw opacity=1 ][fill={rgb, 255:red, 139; green, 180; blue, 229 }  ,fill opacity=1 ][dash pattern={on 4.5pt off 4.5pt}] (341.06,140.33) -- (313.47,140.39) -- (282,140.45) -- (282,79.07) -- (289.47,86.07) -- cycle ;
    \draw  [fill={rgb, 255:red, 113; green, 163; blue, 255 }  ,fill opacity=1 ][dash pattern={on 4.5pt off 4.5pt}] (341.06,140.33) -- (282,201.07) -- (282,140.45) -- cycle ;
    \draw    (341.06,140.33) -- (350,140.42) ;
    \draw [shift={(353,140.45)}, rotate = 180.59] [fill={rgb, 255:red, 0; green, 0; blue, 0 }  ][line width=0.08]  [draw opacity=0] (8.93,-4.29) -- (0,0) -- (8.93,4.29) -- cycle    ;
    
    \draw (131,136.4) node [anchor=north west][inner sep=0.75pt]    {$\boldsymbol{x}$};
    \draw (169,111.4) node [anchor=north west][inner sep=0.75pt]    {$\textcolor[rgb]{0.82,0.01,0.11}{W}$};
    \draw (169,155.4) node [anchor=north west][inner sep=0.75pt]  [color={rgb, 255:red, 208; green, 2; blue, 27 }  ,opacity=1 ]  {$W$};
    \draw (213,100.4) node [anchor=north west][inner sep=0.75pt]    {$\sin( W\cdotp \boldsymbol{x})$};
    \draw (211,165.4) node [anchor=north west][inner sep=0.75pt]    {$\cos( W\cdotp \boldsymbol{x})$};
    \draw (293,111.4) node [anchor=north west][inner sep=0.75pt]  [color={rgb, 255:red, 208; green, 2; blue, 27 }  ,opacity=1 ]  {$\boldsymbol{a}$};
    \draw (293,155.4) node [anchor=north west][inner sep=0.75pt]  [color={rgb, 255:red, 208; green, 2; blue, 27 }  ,opacity=1 ]  {$\boldsymbol{b}$};
    \draw (135,206.04) node [anchor=north west][inner sep=0.75pt]    {$f(\boldsymbol{x} ,\Theta ) =a\cdot \sin( W\cdotp \boldsymbol{x}) +b\cdot \cos( W\cdotp \boldsymbol{x})$};
    \draw (355,136.4) node [anchor=north west][inner sep=0.75pt]    {$f(\boldsymbol{x} ,\Theta )$};
    \draw (129,104.4) node [anchor=north west][inner sep=0.75pt]    {$\mathbb{R}^{d}$};
    \draw (202,59.4) node [anchor=north west][inner sep=0.75pt]    {$\mathbb{R}^{2m} ;\ m\gg 1$};    
    \end{tikzpicture}
\caption{Neural network of 2 layers with Fourier Features model.}
\label{fig:FFmodel}
\end{figure}